\documentclass{ifacconf}

\usepackage{graphicx}      
\usepackage{natbib}        

\usepackage{amsmath,amssymb,bm,url}
\usepackage{algorithm}
\usepackage[noend]{algpseudocode}
\usepackage{etoolbox}
\usepackage{xcolor}
\usepackage{fp, tikz, pgfplots}
\usetikzlibrary{arrows}
\usetikzlibrary{shapes}
\usetikzlibrary{patterns}
\usetikzlibrary{decorations.pathmorphing}
\usetikzlibrary{decorations.pathreplacing}
\usetikzlibrary{decorations.shapes}
\usetikzlibrary{decorations.text}
\usetikzlibrary{shapes.misc}
\usetikzlibrary{decorations.markings}
\usetikzlibrary{arrows.meta}
\usetikzlibrary{backgrounds}
\usepgfplotslibrary{fillbetween}

\algdef{SE}{ParFor}{EndParFor}[1]{\textbf{ParFor} \(\mbox{#1}\) \textbf{do}}{\textbf{end}}%
\makeatletter
\newcommand*{\algrule}[1][\algorithmicindent]{%
  \makebox[#1][l]{%
    \hspace*{.2em}
    \vrule height .75\baselineskip depth .25\baselineskip
  }
}

\pgfplotsset{width=5\columnwidth /5, compat = 1.13, 
	height = 47\columnwidth /100, grid= major, 
	legend cell align = left, ticklabel style = {font=\scriptsize},
	every axis label/.append style={font=\small},
	legend style = {font=\scriptsize},title style={yshift=-7pt, font = \small} }

\newcount\ALG@printindent@tempcnta
\def\ALG@printindent{%
    \ifnum \theALG@nested>0
    \ifx\ALG@text\ALG@x@notext
    \else
    \unskip
    \ALG@printindent@tempcnta=1
    \loop
    \algrule[\csname ALG@ind@\the\ALG@printindent@tempcnta\endcsname]%
    \advance \ALG@printindent@tempcnta 1
    \ifnum \ALG@printindent@tempcnta<\numexpr\theALG@nested+1\relax
    \repeat
    \fi
    \fi
}
\patchcmd{\ALG@doentity}{\noindent\hskip\ALG@tlm}{\ALG@printindent}{}{\errmessage{failed to patch}}
\patchcmd{\ALG@doentity}{\item[]\nointerlineskip}{}{}{} 
\makeatother

\makeatletter
\renewcommand*{\@fnsymbol}[1]{\ensuremath{\ifcase#1\or \or \dagger\or \ddagger\or
		\mathsection\or \mathparagraph\or \|\or **\or \dagger\dagger
		\or \ddagger\ddagger \else\@ctrerr\fi}}
\makeatother

\newcommand\copyrighttext{%
	\footnotesize \textcopyright 2020 the authors. This work has been accepted to IFAC for publication under a Creative Commons Licence CC-BY-NC-ND.}
\newcommand\copyrightnotice{%
	\begin{tikzpicture}[remember picture,overlay]
	\node[anchor=south,yshift=40pt] at (current page.south) {\fbox{\parbox{\dimexpr\textwidth-\fboxsep-\fboxrule\relax}{\copyrighttext}}};
	\end{tikzpicture}%
}

\begin{document}
\begin{frontmatter}

\title{GP3: A Sampling-based Analysis Framework for Gaussian Processes}

\thanks[footnoteinfo]{A. L. gratefully  acknowledges  financial  support from  the German Academic Scholarship Foundation.}
\author[First]{Armin Lederer} 
\author[First]{Markus Kessler} 
\author[First]{Sandra Hirche}

\address[First]{Chair of Information-oriented Control (ITR), Department of
	Electrical and Computer Engineering, Technical University of Munich, Germany (e-mail: {armin.lederer,markus.kessler,hirche}@tum.de).}

\begin{abstract}                
	Although machine learning is increasingly applied in control approaches, only
	few methods guarantee certifiable safety, which is necessary for real world applications. 
	These approaches typically rely on well-understood learning algorithms, which allow 
	formal theoretical analysis. Gaussian process regression is a prominent example among those
	methods, which attracts growing attention due to its strong Bayesian foundations. Even though
	many problems regarding the analysis of Gaussian processes have a similar structure, specific 
	approaches are typically tailored for them individually, without strong focus on computational 
	efficiency. Thereby, the practical applicability and performance of these approaches is 
	limited. In order to overcome this issue, we propose a novel framework called GP3, general 
	purpose computation on graphics processing units for Gaussian processes, which allows to solve
	many of the existing problems efficiently. By employing interval analysis, local Lipschitz 
	constants are computed in order to extend properties verified on a grid to continuous state 
	spaces. Since the computation is completely parallelizable, the computational benefits of 
	GPU processing are exploited in combination with multi-resolution sampling in order to allow 
	high resolution analysis.
\end{abstract}

\begin{keyword}
	Machine learning, Learning for control, Bayesian methods, Gaussian processes, 
	Stability of nonlinear systems, Sampling-based analysis, Learning systems
\end{keyword}

\end{frontmatter}
\copyrightnotice

\setlength{\textfloatsep}{5pt}
\setlength{\abovedisplayskip}{2.2pt}
\setlength{\belowdisplayskip}{2.2pt}

\section{Introduction}

Machine learning is increasingly applied in control approaches, where first principle 
models are not available or expensive to obtain due to the complexity of systems. In 
order to apply learning based control approaches to real world applications, it is 
crucial to certify their safety using theoretically rigorous methods. Although there exists 
a wide variety of machine learning methods, this theoretical analysis is often difficult, 
such that safe control approaches typically rely on a few, well-understood 
learning methods.\looseness=-1

Gaussian process (GP) regression is such a machine learning method which bases on 
solid Bayesian foundations. Due to its inherent bias-variance trade-off it allows efficient
generalization from few training data, which makes it an appealing method both for control
practitioners and theoreticians, and has lead to its increasing use in control. For the control
theoretic analysis of Gaussian processes, several approaches have been developed. In 
\citep{Beckers2016a, Beckers2016} stochastic stability and equilibria of Gaussian processes with 
certain covariance kernels are investigated analytically. Employing numerical quadrature, an 
efficient method to analyze stability of the posterior mean function of a Gaussian process state
space model with squared exponential kernel is developed in \citep{Vinogradska2017a}. A statistical 
learning error analysis of learned MPC control laws is proposed in \citep{Hertneck2018} to ensure 
closed-loop 
stability. By extending the Lyapunov stability conditions on a grid to a continuous state space
using Lipschitz continuity, the region of attraction of a Gaussian process state space model 
is learned in \citep{Berkenkamp2016a}. A similar approach is employed to determine the region of 
attraction of a nonlinear system based on the learned infinite horizon cost function in 
\citep{Lederer2019b}. The Lipschitz constants of the posterior mean function used in these approaches
are also of interest themselves, e.g., for computing uniform regression error bounds \citep{Lederer2019}
or in batch parallelization of Bayesian optimization in \citep{Gonzalez2016}. 

Despite of the similarity of many of these problems, each one is based on a separate analysis method, 
which is typically not optimized for computational efficiency. In order to overcome this issue, 
we propose a novel framework for the analysis of Gaussian process mean functions called GP3: 
\textbf{G}eneral \textbf{P}urpose computation on \textbf{G}raphics \textbf{P}rocessing units for 
\textbf{G}aussian \textbf{P}rocesses. By defining a common problem formulation for many problems, 
we can employ interval analysis to derive local Lipschitz constants on hyperrectangles 
covering the region of interest. These local Lipschitz constants allow to extend verified 
properties on a discrete grid to the whole region of interest such that multi-resolution 
sampling can be used for efficient analysis. As the Lipschitz constants can be computed 
independently for each hyperrectangle, the method is parallelized using general purpose 
graphics processing units in order to exploit the full computational power of modern hardware. 
We demonstrate the flexibility and efficiency of the GP3 framework by applying it to a region
of attraction estimation and a Lipschitz constant bounding problem.

The remaining paper is structured as follows. In Section~\ref{sec:prob} we define the general 
property analysis problem. The theoretical background on Gaussian process regression and 
interval analysis is presented in Section~\ref{sec:theoBack}. The theoretical foundations of the
GP3 framework are explained in Section~\ref{sec:GP_Ana}, before it is evaluated in simulations 
in Section~\ref{sec:numeval}.

\section{Problem Statement}
\label{sec:prob}

Consider the posterior mean function\footnote{Lower/upper case bold symbols denote vectors/matrices, 
	$\mathbb{R}_+$ denotes all real positive numbers,~$\bm{I}_n$ the~$n\times n$ identity matrix and
	$\|\cdot\|$ the Euclidean norm.
	}
$\mu:\mathbb{R}^d\rightarrow\mathbb{R}$ of a Gaussian process with continuous covariance 
kernel \mbox{$k:\mathbb{R}^d\times\mathbb{R}^d\rightarrow\mathbb{R}$}, a continuous function 
$g:\mathbb{R}^d\rightarrow\mathbb{R}$ for comparison and a continuous state transformation 
$\bm{f}:\mathbb{R}^d\rightarrow\mathbb{R}^d$. An abstract problem that finds many practical 
applications is to find bounds~$\epsilon_1,\epsilon_2\in\mathbb{R}_+$ on the difference
\begin{align}
\label{eq:prob}
	-\epsilon_1\leq g(\bm{f}(\bm{x}))-\mu(\bm{x}) \leq \epsilon_2 \quad\forall\bm{x}\in\mathbb{X}
\end{align}
on a compact set~$\mathbb{X}\!\subset\!\mathbb{R}^d$. Various choices for the functions~$\bm{f}(\cdot), g(\cdot)$ immediately come to mind: if we choose~$g(\cdot)$ as the function generating the training data of the Gaussian process and set~$\bm{f}(\bm{x})\!=\!\bm{x}$ we can determine the maximum learning error. Furthermore,~$\bm{g}(\cdot)$ can be the Gaussian process mean~$\mu(\cdot)$ itself and~$\bm{f}(\cdot)$ can be defined such that it returns the closest point in a discrete set. Thereby the variation of the mean function with respect to the discrete set is analyzed. Finally, we can consider~$\bm{f}(\cdot)$ as autonomous, discrete-time dynamics and choose~$g(\cdot)\!=\!\mu(\cdot)$, such that we can immediately investigate if the mean function~$\mu(\cdot)$ satisfies the second condition of Lyapunov's theorem. Although this problem formulation offers such a high flexibility, it allows a straight forward, uniform treatment through interval analysis, which we exploit in the following sections.\looseness=-1

\section{Theoretical Background}
\label{sec:theoBack}

\subsection{Gaussian Process Regression}

Gaussian process regression is a supervised machine learning method, which is frequently applied
in control and system identification due to its Bayesian foundations. The Gaussian process
distribution assigns to any finite subset~$\{ \bm{x}_i,\ldots,\bm{x}_N \}\subset\mathbb{X}$ from
a continuous input domain~$\mathbb{X}\subset\mathbb{R}^d$ a joint Gaussian distribution~\citep{Rasmussen2006}.
It is completely defined through the mean function~$m:\mathbb{R}^d\rightarrow\mathbb{R}$ and 
the covariance function~$k:\mathbb{R}^d\times\mathbb{R}^d\rightarrow\mathbb{R}$. Although prior 
information like approximate models can be incorporated as mean function~$m(\cdot)$ into GP 
regression, such knowledge is often not available, such that the mean function~$m(\cdot)$ is 
usually set to zero. We also assume this in the following. In contrast, a wide variety of 
different covariance functions~$k(\cdot,\cdot)$ is applied in GP regression to encode 
prior information such as smoothness, periodicity and stationarity. Frequently used covariance
functions are the squared exponential kernel 
\begin{align}
k_{\mathrm{SE}}(r)=\sigma_f^2\exp\left( -\frac{1}{2}r^2 \right)
\end{align}
and the Mat\'ern class kernels 
\begin{align}
&k_{m+\frac{1}{2}}(r)=\sigma_f^2p_m(r)\exp\left( -\sqrt{(2m+1)}r \right)
\end{align}
with signal variance~$\sigma_f^2\in\mathbb{R}_+$,~$p_m(\cdot)$ a polynomial of order 
$m\in\mathbb{N}$ and automatic relevance determination distance~\citep{Neal1996}
\begin{align}
\label{eq:autorel}
	r=\sqrt{\sum\limits_{i=1}^d\frac{(x_i-x_i')^2}{l_i^2}}
\end{align}
with length scales~$l_i\in\mathbb{R}_+$. Mat\'ern class kernels are commonly used with 
$m=1$ or~$m=2$ which results in the polynomials
\begin{align}
	p_1(r)&=1+\sqrt{3}r\\
	p_2(r)&=\left(1+\sqrt{5}r+\frac{5}{3}r^2\right).
\end{align}
A major reason for their popularity is the fact that a GP with either of these covariance 
functions is a universal approximator~\citep{Steinwart2001}, i.e., any continuous function 
can be approximated with arbitrary precision.

GP regression is based on the assumption that the training data set 
$\mathbb{D}=\{ (\bm{x}^{(n)},y^{(n)}) \}_{n=1}^N$ is generated through
noisy observations of a function~$h:\mathbb{R}^d\rightarrow\mathbb{R}$, i.e.,
\begin{align}
y^{(i)}= h(\bm{x}^{(i)})+\omega^{(i)},
\end{align}
where~$\omega^{(i)}\sim\mathcal{N}(0,\sigma_n^2)$ are i.i.d. random variables. 
By conditioning the prior joint Gaussian distribution of a prediction~$h(\bm{x})$
and the training outputs~$y^{(i)}$ on the training data~$\mathbb{D}$, we obtain 
the predictive mean 
\begin{align}
\label{eq:mean}
\mu(\bm{x})=\mathbb{E}[h(\bm{x})|\mathbb{D},\bm{x}]=\bm{k}^T(\bm{x})\bm{\lambda},
\end{align}
where 
\begin{align}
\bm{\lambda}=(\bm{K}+\sigma_n^2\bm{I}_N)^{-1}\bm{y}
\end{align}
and the data covariance matrix~$\bm{K}\in\mathbb{R}^{N\times N}$ and the 
covariance vector~$\bm{k}(\bm{x})\in\mathbb{R}^{N}$
are defined through~$K_{ij}=k(\bm{x}^{(i)},\bm{x}^{(j)})$ and 
$k_i(\bm{x})=k(\bm{x}^{(i)},\bm{x})$. The observation noise variance~$\sigma_n^2$, the 
signal variance~$\sigma_f^2$ and the length scales~$l_i$ are considered hyperparameters
of the GP regression and can be determined by maximization of the 
log-likelihood~\citep{Rasmussen2006}.

\subsection{Interval Analysis for Property Analysis}

Interval analysis is a method to approach the problem of calculating bounds of functions. Instead of operating on exact values, interval analysis uses real compact intervals~$[a] = [\underline{a},\bar{a}] = \{x \in \mathbb{R}|\underline{a} \leq x \leq \bar{a}\}$. Basic mathematical operations for intervals are defined as
\begin{align}
	[a] + [b] &= [\underline{a} + \underline{b}, \overline{a} + \overline{b}], \label{eq:IntervalAdd}\\
	[a] - [b] &= [\underline{a} - \overline{b}, \overline{a} - \underline{b}], \label{eq:IntervalSub}\\
	\label{eq:IntervalMul}
	[a] \cdot [b] &= [\mathrm{min}\{\underline{a}\underline{b},\underline{a}\overline{b},\overline{a}\underline{b},\overline{a}\overline{b}\}, \mathrm{max}\{\underline{a}\underline{b},\underline{a}\overline{b},\overline{a}\underline{b},\overline{a}\overline{b}\}].
\end{align}
A more thorough introduction into operations on intervals can be found, e.g., in~\citep{Alefeld2000}. 

Based on this interval arithmetic, it is possible to use intervals as inputs to functions~$f: \mathbb{R} \rightarrow \mathbb{R}$ and calculate output intervals~$[y] = f([x])$, where~$\underline{y}$ serves as a lower bound and~$\bar{y}$ as an upper bound of the function values on the input interval. It is straight forward to adapt this approach to higher dimensional functions~$f: \mathbb{R}^d \rightarrow \mathbb{R}$ by considering so called \textit{hyperrectangles} instead of intervals. A hyperrectangle is completely defined by its center~$\bm{c}$ and length parameter~$\bm{b}$, such that it defines the multi-dimensional interval~$[\bm{c}-\bm{b},\bm{c}+\bm{b}]$ with edge widths~$2\bm{b}$. Using a grid of hyperrectangles, interval analysis allows to efficiently expand the validity of \eqref{eq:prob} on the hyperrectangle centers to the area covered by the hyperrectangles. Using multi-resolution grids as proposed, e.g., in \citep{Bobiti2018}, this enables efficient determination of valid constants~$\epsilon_1$ and~$\epsilon_2$ in \eqref{eq:prob}.

\section{Sampling-Based Analysis of Gaussian Processes}
\label{sec:GP_Ana}

Due to the strong nonlinearity of typical Gaussian process mean functions,
standard interval operations are not directly applicable to their analysis. 
Therefore, we develop an efficient multi-resolution sampling algorithm for 
the analysis of Gaussian processes in Section~\ref{subsec:multi-res}. As this 
algorithm requires upper and lower bounds for the derivative of covariance 
kernels depending on the training data, we investigate such bounds for squared exponential
and Mat\'ern class kernels in Section~\ref{subsec:deriv int}.

\subsection{Multi-resolution Analysis of Gaussian Processes}
\label{subsec:multi-res}

Gaussian processes exhibit a strongly nonlinear mean function in general. In order to efficiently analyze
their mean functions we develop a multi-resolution sampling algorithm in this section. 
Exploiting Lipschitz continuity we derive a theorem to calculate the bounds~$\epsilon_1$ and~$\epsilon_2$ in \eqref{eq:prob} inside hyperrectangles. In order to evaluate the bound, we determine a local Lipschitz constant of the GP on a hyperrectangle. Due to the sampling structure, the analysis of hyperrectangles can be efficiently parallelized using GPUs in each grid refinement iteration.

The basis of our approach lies in the independent analysis of the bounds~$\epsilon_1$ and~$\epsilon_2$ on hyperrectangles. This analysis is founded on the following theorem, which relies on Lipschitz continuity of all involved functions.
\begin{thm}
\label{thmbounds}
	Consider a function~$\bm{f}(\cdot)$ with local Lipschitz constant~$L_f$ and a posterior mean function
	$\mu(\cdot)$ of a Gaussian process with local Lipschitz constant~$L_{\mu}$ on a hyperrectangle
	with center~$\bm{c}$ and edge widths~$2\bm{b}$. Furthermore, assume the function~$g(\cdot)$ has 
	a local Lipschitz constant~$L_g$ on the hyperrectangle with center~$\bm{f}(\bm{c})$ and edge widths 
	$2L_f\bm{b}$. Then, \eqref{eq:prob} holds with 
	\begin{align}
	\label{eq:eps1}
		\epsilon_1&=g(\bm{f}(\bm{c}))-\mu(\bm{c})-(L_fL_g+L_{\mu})\|\bm{b}\|\\
		\epsilon_2&=g(\bm{f}(\bm{c}))-\mu(\bm{c})+(L_fL_g+L_{\mu})\|\bm{b}\|
		\label{eq:eps2}
	\end{align}
	for all~$|\bm{x}-\bm{c}|\leq \bm{b}$, where the 
	absolute values and the comparison are performed element-wise.
\end{thm}
\begin{pf}
	Due to Lipschitz continuity of~$\bm{f}(\cdot)$ and~$g(\cdot)$ it follows that
	\begin{align*}
		|g(\bm{f}(\bm{x}))-g(\bm{f}(\bm{c}))|\leq L_fL_g\|\bm{b}\| \quad\forall\bm{x}:\|\bm{x}-\bm{c}\|\leq \|\bm{b}\|.
	\end{align*}
	Furthermore, Lipschitz continuity of the GP yields
	\begin{align*}
		|\mu(\bm{x})-\mu(\bm{c})|\leq L_{\mu}\|\bm{b}\|.
	\end{align*}
	Applying the triangle inequality we finally obtain the bounds \eqref{eq:eps1} and \eqref{eq:eps2}.
\end{pf}

The application of this theorem crucially relies on the Lipschitz constant of the GP mean function. The computation of this Lipschitz constant on a hyperrectangle can be performed independently for each hyperrectangle and depends merely on the parameter vector~$\bm{\lambda}$, the kernel function~$k(\cdot,\cdot)$ and the data points~$\bm{x}^{(i)}$,~$i=1,\ldots,N$, as shown in the following theorem.

\begin{thm}
\label{thmlipconst}
	Consider the posterior mean function~$\mu(\cdot)=\bm{k}^T(\cdot)\bm{\lambda}$
	of a GP with covariance kernel~$k(\cdot,\cdot)$ on a hyperrectangle with 
	center~$\bm{x}$ and length~$2\bm{b}$. Let~$\bm{L}_{k,i}^{\partial j}$ denote 
	a vector of partial derivative bounds of the~$i$-th element of~$\bm{k}(\cdot)$ 
	with respect to~$x_j$. Then, a local Lipschitz constant of the mean on the 
	hyperrectangle is given by
	\begin{align}
		L_{\mu}(\bm{x},\bm{b})=\sqrt{\sum\limits_{j=1}^d \max\left\{ \left(\sum\limits_{i=1}^N\bm{R}(\lambda_i)\lambda_i\bm{L}_{k,i}^{\partial j} \right)^2 \right\}},
	\end{align}
	where
	\begin{align}
		\bm{R}(\lambda_i)=\begin{cases}
		\bm{I}_2  & \lambda_i>0\\
		\begin{bmatrix}
		0&1\\1&0
		\end{bmatrix}&\lambda_i\leq 0.
		\end{cases}
	\end{align}
\end{thm}
\begin{pf}
	Due to the scalar product of parameter vector~$\bm{\lambda}$ and the kernel vector~$\bm{k}(\bm{x})$ in \eqref{eq:mean}, the partial derivatives of the mean function are given by
	\begin{align*}
		\frac{\partial}{\partial x_j}\mu(\bm{x})=\begin{bmatrix}
		\frac{\partial}{\partial x_j} k_1(\bm{x}) & \ldots & \frac{\partial}{\partial x_j} k_N(\bm{x})
		\end{bmatrix}\bm{\lambda}.
	\end{align*} 
	The vectors~$\bm{L}_{k,i}^{\partial j}$ contain upper and lower bound on the partial 
	derivatives~$\frac{\partial}{\partial x_j} k_i(\bm{x})$ in the first and second row, respectively. 
	By multiplying these vectors with the matrix~$\bm{R}(\lambda_i)$, the order of elements is 
	changed if~$\lambda_i$ is negative, such that upper derivative bounds are multiplied with 
	positive~$\lambda_i$s and lower bounds with negative ones in the first row of the resulting vector. 
	The multiplication is performed in the inverse combination in the second row. By summing up each 
	row, upper and lower bounds on the partial derivatives are obtained. We take the maximum squared 
	value of these two rows as squared Lipschitz constant in the~$j$-th direction and finally, 
	calculate the overall Lipschitz~$L_{\mu}(\bm{x},\bm{b})$ constant by taking the square root of 
	the sum of squared Lipschitz constants in all directions.\looseness=-1
\end{pf}

Theorem \ref{thmbounds} and Theorem \ref{thmlipconst} allow for a straightforward implementation and integration in a multi-resolution, GPU parallelized sampling algorithm, which is depicted in Algorithm \ref{Alg1}. Initially, a sampling grid is created by dividing the analyzed region~$\mathbb{X}$ into~$M$ hyperrectangles in line 2, where the centers~$\bm{c}_i$ are concatenated in~$\bm{C}$ and edge-lengths~$\bm{b}_i$ are concatenated in~$\bm{B}$. Generally, any initial sampled grid is possible for this step, since our approach does not crucially depend on it. The grid is refined in line 6 by using an indication value~$s_i$, which divides hyperrectangles that must be further refined for~$s_i = \texttt{false}$ from those that require no further refinement. For the refinement procedure itself, several ways are possible in general. A simple method is to resample the grid with smaller hyperrectangle sizes~$\bm{B}$ and skip the calculation where~$s_i=\texttt{true}$ already. Another example for a refinement method is the~$2$-refinement used in~\citep{Bobiti2018}, where hyperrectangles with~$s_i=\texttt{false}$ are divided into~$2^d$ smaller hyperrectangles with new edge widths~$\frac{1}{2}\bm{b}_i$. \looseness=-1

The computation of the bounds~$\epsilon_1,\epsilon_2$ based on Theorems~\ref{thmbounds} and~\ref{thmlipconst} takes place inside the parallelized For (\textbf{ParFor}) loop of line 7. By shifting the execution to a highly parallel GPU, the computation time of the algorithm can be significantly reduced, which is an important feature of this approach. The termination of the refinement process of the algorithm is ensured by setting~$s_i \!=\! \texttt{true}$, once the bounds~$\epsilon_1,\epsilon_2$ inside the hyperrectangle satisfy predefined desired bounding functions~$\bar{\epsilon}_1(\bm{c}_i)$ and~$\bar{\epsilon}_2(\bm{c}_i)$ or the size~$\|\bm{b}_i\|$ of the hyperrectangle falls below a specified minimum size~$b_{\mathrm{min}}$.\looseness=-1
\begin{algorithm}[t!]
\caption{Multi-resolution bound calculation}
\label{Alg1}
\begin{algorithmic}[1]
\Function{Bounds}{$g(\cdot),\mu(\cdot),\mathbb{X},M,\bar{\epsilon}_1(\cdot),\bar{\epsilon}_2(\cdot),b_{\mathrm{min}}$}
  \State~$[\bm{C};\bm{B}] \gets$ \Call{SamplingGrid}{$\mathbb{X},M$} 
  \State~$\bm{\epsilon}_1,\bm{\epsilon}_2 \gets \texttt{zeros}[\textsf{len}(\bm{C})]$
  \State~$\bm{s}\gets\texttt{false}[\textsf{len}(\bm{C})]$
  \While{$\textsf{any}(s_i) = \texttt{false}$} 
  	\State~$[\bm{C};\bm{B}] \gets$ \Call{RefineGrid}{$\bm{C};\bm{B};s_i$} 
  	\ParFor{$i=1$ to len$(\bm{C})$} 
  		\State Calculate~$L_{\mu}(\bm{c}_i, \bm{b}_i)$ using Theorem~\ref{thmlipconst}
  		\State Calculate~$\epsilon_{1,i},\epsilon_{2,i}$ based on Theorem~\ref{thmbounds}
  		\State$s_i \gets (\epsilon_{1,i}\leq\bar{\epsilon}_1(\bm{c}_i) \land \epsilon_{2,i}\leq\bar{\epsilon}_2(\bm{c}_i)) \lor$
  		\State\hphantom{$s_i \gets\ $}$(\|\bm{b}_i\| \leq b_{\mathrm{min}})$
  		\EndParFor
  \EndWhile
\State \Return~$\bm{C},\bm{\epsilon}_1,\bm{\epsilon}_2$
\EndFunction
\end{algorithmic}
\end{algorithm}

\subsection{Derivative Intervals of Kernels}
\label{subsec:deriv int}

Typical covariance kernels are highly nonlinear functions themselves, which
complicates the interval analysis of their derivatives. However, many kernels 
exhibit a structure such that derivatives are monotonous on large intervals. 
We exploit this behavior for deriving upper and lower bounds for squared 
exponential and Mat\'ern class kernels in the following theorem.
\begin{thm}
	Consider a multivariate squared exponential or Mat\'ern class
	kernel~$k(\cdot)$ with length scale~$\bm{l}$
	on a hyperrectangle with center~$\bm{x}$ and edge lengths~$2\bm{b}$.
	Then, the derivative bounds for a training point~$\bm{x}^{(i)}$ with 
	respect to the~$j$-th component is given by
	\begin{align}
	\label{eq:boundsmult}
	\bm{L}_{k,i}^{\partial j}=\bm{T}(x_j^{(i)}-x_j)\bm{\kappa}(\bm{x}^{(i)}-\bm{x})	
	\end{align}
	where
	\begin{align}
	&\bm{\kappa}(\bm{\Delta})=\\
	&\begin{cases}
	\begin{bmatrix}
	-\frac{\partial }{\partial \Delta_j}k\left(\rho_j\left(\Delta-\overline{\bm{\tau}},|\Delta_j|-b_j\right)\right)\\
	-\frac{\partial }{\partial \Delta_j}k\left(\rho_j\left(\Delta-\underline{\bm{\tau}},|\Delta_j|+b_j\right)\right)
	\end{bmatrix}& |\Delta_j|>b_j+\tilde{l}_j\\[18pt]
	\begin{bmatrix}
	-\frac{\partial }{\partial \Delta_j}k\left(\rho_j\left(\Delta-\overline{\bm{\tau}},|\Delta_j|+b_j\right)\right)\\
	-\frac{\partial }{\partial \Delta_j}k\left(\rho_j\left(\Delta-\underline{\bm{\tau}},|\Delta_j|-b_j\right)\right)
	\end{bmatrix}& |\Delta_j|<\tilde{l}_j-b_j\\[18pt]
	\begin{bmatrix}
	-\frac{\partial }{\partial \Delta_j}k\left(\rho_j\left(\Delta-\overline{\bm{\tau}},\tilde{l}_j\right)\right)\\
	\!\min\left\{\! -\frac{\partial }{\partial \Delta_j}k\left(\rho_j\left(\Delta\!-\!\underline{\bm{\tau}},|\Delta_j|\!\pm\! b_j\right)\right) \!\right\}\!
	\end{bmatrix}\!&\!||\Delta_j|\!-\!\tilde{l}_j|\leq b_j
	\end{cases}
	\label{eq:kappa}\\
	&\bm{T}(\Delta_j)=\begin{cases}
	\bm{I}_2&\Delta_j>0\\
	\begin{bmatrix}
	0&-1\\-1&0
	\end{bmatrix}&\Delta_j\leq 0
	\end{cases},
	\end{align}
	with modified weighted distance~$\rho_j(\bm{\Delta},\tilde{\Delta})$, maximum point~$\tilde{l}_j$,
	maximum distance~$\overline{\bm{\tau}}$, minimum distance~$\underline{\bm{\tau}}$ and index 
	set~$\mathbb{I}_j$ defined as follows
	\allowdisplaybreaks
	\begin{align}
	\rho_j(\!\bm{\Delta},\tilde{\Delta}\!)&=\sqrt{\!\frac{\tilde{\Delta}^2}{l_j^2}\!+\!\sum\limits_{i\in\mathbb{I}_j}\!
		\frac{\Delta_i^2}{l_i^2}\!}\\
	\tilde{l}_j&=\begin{cases}	
	l_j\!& \!\text{squared exponential kernel}\\
	\frac{l_j}{\sqrt{3}}\!&\!\text{Mat\'ern kernel with }m=1\\
	\frac{5\!+\!\sqrt{5}}{10}l_j\!&\!\text{Mat\'ern kernel with }m=2
	\end{cases}\\
	\label{eq:taubar}
	\overline{\bm{\tau}}&= \min\left\{ \|\bm{b}_{\neq j}\|, \|\bm{x}_{\neq j}^{(i)}\!-\!\bm{x}_{\neq j} \|\right\}
	\frac{\bm{x}^{(i)}\!-\!\bm{x}}{\|\bm{x}_{\neq j}^{(i)}\!-\!\bm{x}_{\neq j} \|}\\
	\label{eq:tauunder}
	\underline{\bm{\tau}}&=\|\bm{b}_{\neq j}\|\frac{\bm{x}\!-\!\bm{x}^{(i)}}{\|\bm{x}_{\neq j}^{(i)}\!-\!\bm{x}_{\neq j} \|}\\
	\mathbb{I}_j&=\{1,\ldots,j-1,j+1,\ldots,d\}.
	\end{align}
\end{thm}
\begin{pf}
	We start this proof by first showing that \eqref{eq:boundsmult} holds for
	the squared exponential before we highlight the differences in the proof 
	for Mat\'ern kernels. Due to the exponent product rule and the automatic relevance
	determination \eqref{eq:autorel}, the derivative of 
	the multivariate squared exponential kernel can be split into
	\begin{align*}
	\frac{\partial }{\partial x_j}k(\bm{x}^{(i)},\bm{x})=-\frac{\partial }{\partial \Delta_j}k\left(\frac{|\Delta_j|}{l_j}\right)k\left(\sqrt{\sum\limits_{i\in\mathbb{I}_j} \frac{\Delta_j^2}{l_i^2}}\right),
	\end{align*}
	where~$\bm{\Delta}=\bm{x}^{(i)}-\bm{x}$. This separation allows to determine interval
	bounds for both factors independently and combine the maximizers or minimizers, respectively, 
	with the help of~$\rho_j(\bm{\Delta},\Delta_j)$ based on \eqref{eq:IntervalMul}. Due to odd 
	symmetry of the univariate derivative, it is sufficient to derive only bounds for the positive 
	real line and obtain the interval bounds for the negative real line by multiplication with 
	$\bm{T}(\Delta_j)$. The kernel derivative is monotonous on the intervals~$[0,l_j]$ and~$[l_j,\infty]$ 
	with a maximum at~$l_j$. Therefore, we obtain three intervals with maximizers~$\nu$
	\begin{align*}
		|\Delta_j|>b_j+l_j&: \nu=|\Delta_j|-b_j\\
		|\Delta_j|<l_j-b_j&: \nu=|\Delta_j|+b_j\\
		||\Delta_j|-l_j|\leq b_j&: \nu=l_j
	\end{align*}
	which can be analogously obtained for the minimizers. 
	The multivariate kernel is trivially maximized by considering the minimal distance to the
	training point, which is given by~$\bm{\Delta}-\overline{\bm{\tau}}$, while 
	it is minimized by~$\bm{\Delta}-\underline{\bm{\tau}}$. Therefore, \eqref{eq:boundsmult}
	provides an upper and lower bound for the derivative of the squared exponential kernel 
	on the hyperrectangle and we proceed with Mat\'ern class kernels. Although the derivatives
	of Mat\'ern class kernels cannot be separated as the squared exponential kernel, they 
	exhibit a similar behavior in that they are monotonous in the non-derived directions 
	$i\neq j$. Therefore, we can choose the same maximizer and minimizer in these directions
	as for the squared exponential kernel. Furthermore, the derivatives of Mat\'ern class 
	kernels exhibit the same odd symmetry in the derived direction~$j$ and have two monotonous
	intervals on the positive real line with maximum at~$\tilde{l}_j=l_j/\sqrt{3}$ for 
	$m=1$ and~$\tilde{l}_j=(5+\sqrt{5})l_j/10$. Therefore, we can upper and 
	lower bound the derivative for Mat\'ern class kernels on the positive real line using 
	\eqref{eq:kappa}, which concludes the proof.\looseness=-1
\end{pf}
Although this theorem might appear rather complicated, it allows a straightforward 
implementation with few conditional operators, which is beneficial for GPU parallelization~\citep{GPUcomputing}. 
Therefore, it allows efficient, parallelized analysis of Gaussian process mean functions 
in combination with Theorems~\ref{thmbounds} and~\ref{thmlipconst}.

\section{Numerical Evaluation}
\label{sec:numeval}

\subsection{Efficiency of the GP3 Framework}
\label{subsec:effic}
In order to demonstrate the advantages of the GP3 framework\footnote{Code is available at \url{https://gitlab.lrz.de/alederer/gp3}} 
over a global Lipschitz 
constant computation and a CPU parallelized implementation, we compare these approaches 
on the example proposed in~\citep{Lederer2019}. A crucial step therein is the derivation of
a Lipschitz constant for a posterior mean representing the function\looseness=-1
\begin{align}
	f(\bm{x})=1-\sin(x_1)+\frac{1}{1+\exp(-x_2)}.
\end{align}
The Gaussian process is trained with~$100$ samples which are uniformly spaced 
over the analyzed region 
$\mathbb{X}=[-6,4]\times[-4,4]$. 
We train three GPs with a squared exponential and
Mat\'ern kernels with~$m=1$ and~$m=2$. The hyperparameters
obtained via log-likelihood maximization for each of these GPs are depicted in 
Table~\ref{tab:hyp_param}.
\begin{table}[b]
	\centering
	\begin{tabular}{l| c c c}
		kernel &~$\sigma_n^2$ &~$\sigma_f^2$ &~$\bm{l}$\\ \hline 
		squared exponential &~$0.1$ &~$0.956$ &~$[1.762\ 5.537]^T$\\
		Mat\'ern with~$m=1$ &~$0.1$ &~$1.274$ &~$[3.755\ 15.052]^T$\\
		Mat\'ern with~$m=2$ &~$0.1$ &~$1.012$ &~$[2.333\ 8.496]^T$
		\rule{0pt}{3ex}   
	\end{tabular}
	\caption{Hyperparameters of the different covariance kernels}
	\label{tab:hyp_param}
\end{table}

\begin{figure}
	\centering
	\includegraphics[]{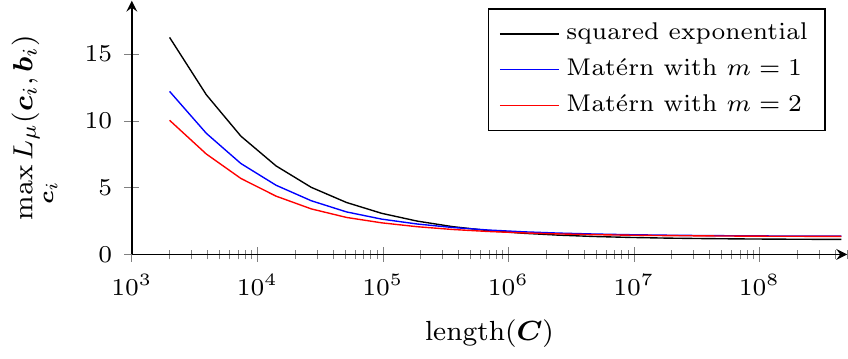}
	\vspace{-0.3cm}
	\caption{Decrease rate of Lipschitz constants for increasing number of analyzed hyperrectangles}
	\label{fig:Lip}
\end{figure}

We compute the Lipschitz constants of the posterior mean functions using 
the GP3 framework for different numbers of hyperrectangles and the naive, global 
Lipschitz constant proposed in~\citep{Lederer2019}. The Lipschitz constants obtained from 
the GP3 approach are depicted in Figure~\ref{fig:Lip}. Since the naive approach yields 
Lipschitz constants of~$343.62$,~$245.66$ and~$268.20$ for the squared exponential and 
Mat\'ern kernel with~$m=1$ and~$m=2$, respectively, 
the corresponding constant curves are not displayed in the figure. In contrast, the 
Lipschitz constants obtained by the GP3 approach can provide Lipschitz constants 
smaller than~$20$ with less than~$2000$ hyperrectangles. Furthermore, with 
approximately~$10^6$ hyperrectangles the Lipschitz constant has almost converged to 
a constant value.\looseness=-1

\begin{figure}
	\centering
	\includegraphics[]{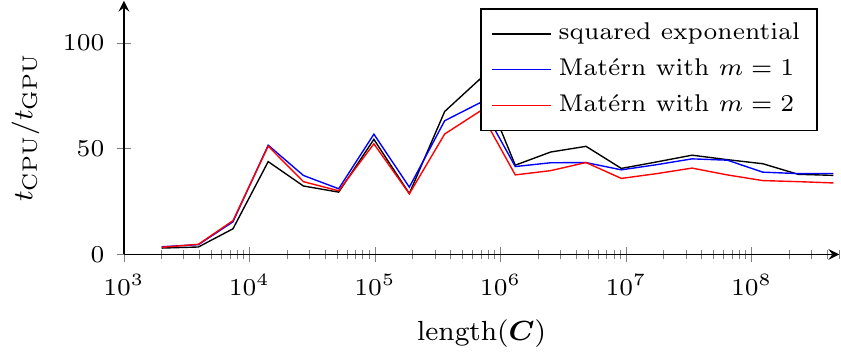}
	\vspace{-0.3cm}
	\caption{Average speedup of GPU over CPU parallelization over the number of analyzed hyperrectangles}
	\label{fig:comptime}
\end{figure}

Additionally, we compare the time required to compute the Lipschitz constants on the 
hyperrectangles with a GPU and CPU parallelization. We compute the Lipschitz constants 
on a system with a NVIDIA TITAN V GPU, which has~$2\!\!:\!\!1$ single to double precision unit 
ratio, two AMD EPYC~$16$-Core CPUs and~$1$TB RAM. The speedup of the GPU
compared to the CPU parallelization averaged over~$10$ runs of both 
implementations are displayed for the three different covariance kernels in 
Figure~\ref{fig:comptime}. Although we parallelize the Lipschitz constant computation 
with~$64$ threads on the CPU, the GPU parallelization achieves a speedup
of at least~$30$ for large numbers of hyperrectangles. Merely at low numbers of hyperrectangles 
the speedup can be in the single digit region due to the computational overhead of GPU computation.
Therefore, GP3 allows to exploit the advantages of modern general purpose GPU computing to achieve 
low computation times, while requiring a reasonable amount of hyperrectangles for convergence of the 
obtained Lipschitz constant.

\subsection{Region of Attraction for Power Systems}

As an example for a problem of the form \eqref{eq:prob}, we analyze the region of attraction 
of a nonlinear autonomous system. We consider the single machine infinite bus
system~\citep{Munz2013}
\begin{align}
\label{eq:smibs}
m_1\ddot{\phi}_1+d_1\dot{\phi}_1=-a_{12}(\sin(\theta_1+\phi_1)-\sin(\theta_1)),
\end{align}
which models a synchronous machine with inertia~$m_1$, damping~$d_1=20$ and steady state 
phase~$\theta_1=\arcsin(0.05)$ as generator bus connected to an infinite bus with~$\theta_2=\phi_2=0$, 
such that~$a_{12}=10$ is the product of the susceptance between both buses and the root mean square 
voltages~$u_1$ and~$u_2$ at bus~$1$ and~$2$, respectively. For analyzing the regions, where the system
is discrete-time stable, we consider the cost of finite time trajectories proposed in \citep{Bobiti2018} 
as Lyapunov function, i.e., 
\begin{align}
\label{eq:Lyap}
	V(\bm{x}_0)=\sum\limits_{k=0}^K\bm{x}^2(k\Delta t,\bm{x}_0)
\end{align}
where~$\bm{x}(t,\bm{x}_0)\!=\![\dot{\phi}(t)\ \phi(t)]$ denotes the solution of the 
differential equation \eqref{eq:smibs} for initial state~$\bm{x}_0$ and
\linebreak~$\Delta t\!\in\!\mathbb{R}_+$ is the sampling time. For determining the states at 
the sampling times~$k\Delta t$, we numerically
integrate the system using the Bogacki-Shampine method \citep{Bogacki1989}. In order to avoid 
the computational complexity of performing this numerical integration for many test points, 
we compute the value of \eqref{eq:Lyap} with~$K\!=\!1000$ and~$\Delta t\!=\!0.01$ only for~$N\!=\!1024$ initial states 
$\bm{x}_0$ uniformly spread over 
the rectangle~$\mathbb{X}\!=\![-5,5]^2$ and train a Gaussian process with the data to obtain a learned Lyapunov
function~$\tilde{V}(\cdot)\!=\!\mu(\cdot)$.\looseness=-1

In order to analyze the region of attraction of the system, we follow the approach proposed in 
\citep{Lederer2019b}. First, we determine the regions~$\mathbb{W}$ 
of the state space satisfying the inequality
\begin{align}
	\Delta V(\bm{x}_0)=\tilde{V}(\bm{x}(\Delta t,\bm{x}_0))-\tilde{V}(\bm{x}_0)\leq 0,
\end{align}
which corresponds to verifying \eqref{eq:prob} with~$\epsilon_1=\infty$,~$\epsilon_2=0$, 
$g(\cdot)=\mu(\cdot)$ and 
$\bm{f}(\cdot)=\bm{x}(\Delta t,\cdot)$. Based on the decrease region~$\mathbb{W}$, the region of attraction 
is a level set of the learned Lyapunov given by
\begin{align}
	\mathbb{V}=\{\bm{x}\in\mathbb{X}:\tilde{V}(\bm{x})\leq \min\limits_{\bm{x}\in\mathbb{X}: \Delta \tilde{V}(\bm{x})>0} \tilde{V}(\bm{x})\},
\end{align}
which can also be determined using the GP3 framework. Since the Lipschitz based analysis of 
our approach does not allow the verification of the decrease condition close to the origin, we assume
stability in a ball around~$\bm{0}$ with radius~$r=0.1$ similarly as in \citep{Lederer2019b}.\looseness=-1

\begin{figure}
	\includegraphics[]{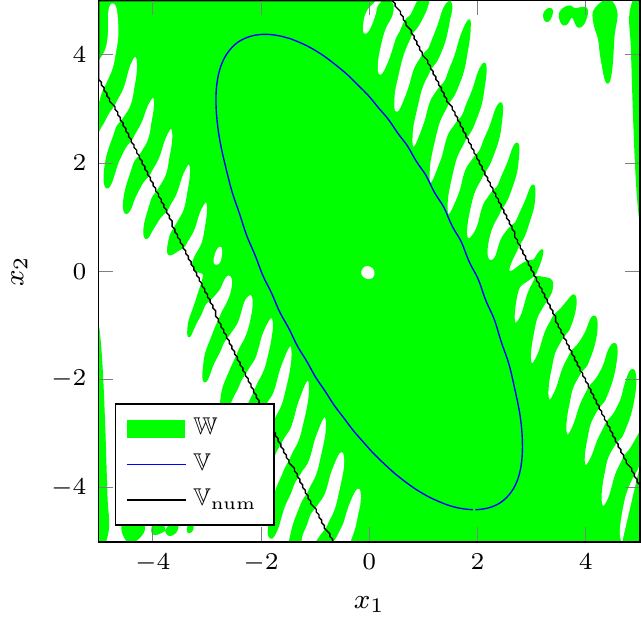}
	\vspace{-0.3cm}
	\caption{Lyapunov decrease region~$\mathbb{W}$ and region of attraction~$\mathbb{V}$ 
		obtained with Alg.~\ref{Alg1} as well as numerically approximated region of 
		attraction~$\mathbb{V}_{\mathrm{num}}$ for the single machine infinite bus system}
	\label{fig:roa}
\end{figure}

We apply Alg.~\ref{Alg1} with minimal size~$b_{\min}=10^{-9}$ of the hyperrectangles 
and a conservative Lipschitz constant~$L_f=20$ to both problems. 
The resulting decrease region~$\mathbb{W}$ as well as region of attraction~$\mathbb{V}$ of the 
single machine infinite bus system are illustrated in Fig.~\ref{fig:roa}. Additionally, an estimate 
of the region of attraction~$\mathbb{V}_{\mathrm{num}}$ obtained through examining the convergence of 
trajectories after~$10^4$ simulation steps for~$4000$ initial states~$\bm{x}_0$ is depicted. The 
verified decrease region is 
different from the approximated true region of attraction~$\mathbb{V}_{\mathrm{num}}$ at its boundary. 
Due to the saw tooth behavior at the boundary of the verified decrease region~$\mathbb{W}$, the resulting
region of attraction~$\mathbb{V}$ is smaller than the numerical approximation~$\mathbb{V}_{\mathrm{num}}$. 
However, this underestimation results mainly from imprecision of the learning, while the GP3 approach allows to 
analyze the Gaussian process very accurately, since the boundary of~$\mathbb{V}$ visually
touches the boundary of~$\mathbb{W}$ at~$\bm{x}=[1.2\ 1.7]^T$. Therefore, better estimates of 
the region of attraction can easily be obtained by training the Gaussian process with more data of the 
discrete-time Lyapuonv function \eqref{eq:Lyap}.

\section{Conclusion}
\label{sec:conc}

This paper introduces a novel framework for the analysis of Gaussian process mean 
functions called GP3: general purpose computation on graphics processing units for
Gaussian processes. Based on interval analysis to compute local Lipschitz constants, 
the posterior mean function is analyzed using multi-resolution sampling. Due independence
of the computations for each sample, the method can be parallelized on a GPU for computational
efficiency. In order to demonstrate the computational benefits of the GP3 framework, it is 
applied to a Lipschitz constant bounding and a region of attraction estimation problem.

\bibliography{ifacconf}             
\end{document}